\documentclass[a4paper]{article}
\usepackage[english]{babel}

\usepackage{amsmath}
\usepackage{graphicx}
\usepackage{etoolbox}
\usepackage{changepage}
\usepackage{titlesec}
\usepackage[parfill]{parskip}
\usepackage[margin=1in]{geometry}
\usepackage{times}
\usepackage{float}
\usepackage{subcaption} 
\usepackage[backend=biber,sorting=ynt]{biblatex}
\usepackage{hyperref}  
\usepackage{authblk}
\usepackage[utf8]{inputenc}
\providecommand{\keywords}[1]{\textbf{\textit{Keywords:}} #1}

\title{\large \textbf{Stacking with Neural network for Cryptocurrency investment}} 

\author[1]{Avinash Barnwal}
\author[1]{Hari Pad Bharti}
\author[1]{Aasim Ali}
\author[1]{Vishal Singh}
\affil[1]{Inncretech Inc., Princeton}
\affil[ ]{\textit {\{avinash,hari,aasim,vishal\}@inncretech.com}}

\makeatletter 
\patchcmd{\@maketitle}{\begin{center}}{\begin{adjustwidth}{0.5in}{0.5in}\begin{center}}{}{}
\patchcmd{\@maketitle}{\end{center}}{\end{center}\end{adjustwidth}}{}{}
\makeatother

\begin{document}
\raggedright
\maketitle
\thispagestyle{empty}
\pagestyle{empty}

\begin{abstract}
    Predicting the direction of assets have been an active area of study and difficult task. Machine learning models have been used to build robust models to model the above task. Ensemble methods are one of them resulting better than single supervised method. In this paper, we have used generative and discriminative classifiers to create the stack,particularly 3 generative and 6 discriminative classifiers and optimized over one-layer Neural Network to model the direction of price cryptocurrencies. Features used are  technical indicators used are not limited to trend, momentum, volume, volatility indicators and sentiment analysis has also been used to gain useful insight combined with above features. For Cross validation, Purged Walk forward cross validation has been used. In terms of accuracy, we have done comparative analysis of the performance of Ensemble method with Stacking and Ensemble method with blending. We have also developed methodology for combined features importance for stacked model. Important indicators are also identified based on feature importance. 
\end{abstract}


\begin{flushleft}
\footnote{Permission to make digital or hard copies of part or all of this work for personal or classroom use is granted without fee provided that copies are not made or distributed for profit or commercial advantage and that copies bear this notice and the full citation on the first page. Copyrights for third-party components of this work must be honored.
For all other uses, contact the owner/author(s).}
\end{flushleft}


\section{\Large{Introduction}}
\setcounter{section}{1}

Today , there more are more than 1000 cryptocurrencies. Having
nearly \$200 Billion of market capitalization and daily volume of nearly \$15 Billion. Bitcoin,Ethereum,Ripple,Bitcoin Cash and Stellar being top 5 cryptocurrencies based on market capitalization.Previous studies include price formation of Bitcoin and identifying important features to drive the price \cite{1}.

The Crash of cryptocurrencies in 2018 made it evident that it is complex , dynamic and non-linear. The behavior is not very different compared to stock markets where sharp rise in measures of collective behavior was observed \cite{2}. Assets direction predictability has been one of key area of study for portfolio management. Being complex,dynamic and non linear in nature makes it more difficult to develop robust strategies. Many authors have attempted to develop machine learning models for financial trading and success of it for Stock market prediction makes it suitable for cryptocurrencies price direction prediction.

Deep Learning has been applied for forecasting stock returns \cite{3,4}. It is shown that it is more successful than shallow neural networks. There are also other machine learning models applications which have shown great performance like Gradient Boosting \cite{5}, Bayesian Neural network \cite{6}, LSTM \cite{7}, Naive bayes \cite{8}, Random Forest \cite{7} and many more.
\cite{9} talks about predicting stock market movement direction with SVM combined with LDA, QDA, and Neural network but doesn't address through perspective of discriminative and generative models \cite{10}. There are some trade offs of using each models.

Combining different models can lead to better prediction result \cite{11} but there are two ways to combine it blending and stacking. Stacked Generalization \cite{12} introduces the concept of meta-learners. It combines different models unlike bagging and boosting. With new machine learning models developed like xgboost \cite{13} and LightGBM \cite{14}, diversed base learners are included.

Technical indicators are mainly used with other fundamental indicators to develop trading strategies or models to prediction prices. As cryptocurrencies are used, fundamental indicators are not included. Along with Technical indicators , sentiment indicators are also used.Tweets by coindesk are used for creating sentiment indicators. Coindesk is considered to be leading news provider for Blockchain. Twitter data can be used to analyse the investor sentiment and leading to price formation of stocks and Bitcoin.

The main contributions are: 
(1) We formulate the problem of predicting direction of bitcoin's price. 
(2) We have created the feautures using Technical Indicators including Momentum,Trend,Volume and Volatility and Sentiment Indicators using Tweets by Coindesk. 
(3) Mixing Discriminative and Generative models to create class of base learners including non-linear models to capture non-separability in the models. 
(4)Tuning hyper parameters of models using Pursed Time Series Cross-Validation to estimate robust models.
(5)Improving model performance using Stacking of base learners where Stacking model is 1-Layer Feed Forward Neural network.
(6)Finding important features using Partial Dependence Plot which is important to day-traders. 

The remainder of paper is structured as follows.In section Materials and Methods , we describe the data,indicators,Comparison between Discriminative and Generative Models, Models, Cross-Validation Technique and Stacking using 1-Layer Neural Network. In section Results, we present the hyper parameters tuning of different models and corresponding performance of each model in terms of log-loss, Accuracy, Recall and F1-Score. Feature Importance is calculated using partial-dependence plot for each model and also developed methodology to calculate feature importance for stacked model. In section Conclusion, we conclude and highlight the key results. 


\section{\Large{Materials and Methods}}
\setcounter{section}{2}

\subsection{\large{Data description and pre-processing}}

Bitcoin data is downloaded from quandl. Quandl offers to download bitcoin data from different exchanges to capture the true price of digital assets. We have considered four exchanges KRAKEN, BITSTAMP, ITBITUSD and COINBASE to remove ambiguity and final price is created based on weighted volume
price. Missing data in price is imputed with exponential average technique. We have considered time period from Aug-2017 to Jul-2018 with end of the day data. This period covers the peak time and down time as well. Therefore this will be a right time to test the strategy as it includes bull period and bear period.

Data dredging or cherry pick are one of pitfalls of back-testing any strategy which we have avoided by picking diversed time period and testing only price direction prediction.

Next we have technical indicators as features in the model.We have considered four types of technical indicators Volume, Volatility, Trend and Momentum.

\textbf{Technical Indicators}\\
\vspace{2mm}
Notations used -  \\
$high_t$ - Highest price for that day \\
$low_t$  - Lowest price for that day \\
$close_t$ - Close price for that day \\
$open_t$ - Open price for that day \\
$volume_t$ - Volume for that day
$EMA(X,n)$ - Exponential Moving Average of X with window n
$MA(X,n)$  - Moving Average of X with window n
\vspace{2mm}

\begin{tabular}{ |p{3cm}||p{3cm}|p{3cm}|p{3cm}|  }
 \hline
 \multicolumn{4}{|c|}{Technical Indicators} \\
 \hline
 Volume& Volatility& Trend& Momentum\\
 \hline
 Accumulation Distribution Index(ADI)&Average \ true range(ATR)&Moving Average Convergence Divergence(MACD)&   Relative Strength Index (RSI)\\
 On balance volume& Bollinger Moving Average  & Moving Average Convergence Divergence Signal   &True strength index (TSI)\\
 On balance volume mean& Bollinger Lower Band & Moving Average Convergence Divergence Diff&  Stochastic Oscillator\\
 Chaikin money flow&Bollinger Higher Band & Exponential Moving Average&  Williams \%R\\
 Force index&   Bollinger Higher Band Indicator  & Vortex Indicator Negative (VI)&Awesome Oscillator\\
 Volume Price Trend& Bollinger Lower Band Indicator  & Trix (TRIX)   & \\
 Negative volume index& Keltner Channel Central  & Mass Index(MI)& \\
    & Keltner Channel Higher Band& Commodity Channel Index (CCI)& \\
    & Keltner Channel Lower Band& Detrended Price Oscillator (DPO)& \\
    & Keltner Channel Higher Band Indicator& KST Oscillator (KST)& \\
    & Keltner Channel Lower Band Indicator& KST Oscillator (KST Signal)& \\
    & Donchian Channel Higher Band&  & \\
    & Donchian Channel Lower Band&  & \\  
    & Donchian Channel Higher Band Indicator&  & \\  
    & Donchian Channel Lower Band Indicator&   & \\  
    
 \hline
\end{tabular}

Next, we have used tweets to create the sentiment indicator.

\subsection{\large{Model}}
Lets say we have feature vector $X_t$ $\epsilon$ $R^n$ to build the model having dependent variable $y_{t+1}$ , here
$y_{t+1}$ is defined based on the return of the asset $r_{t+1}$ where 

\begin{equation}
y_{t+1} = 
\begin{cases}
    1, & r_{t+1} > 0\\
    0 & r_{t+1} \leq 0\\
\end{cases}
\end{equation}

Building robust model involving time series data set can be difficult as nature of the asset can vary a lot from one time period to another. Therefore its important to consider wide range of predictive models to capture the linear and non linear relationship between feature vector and dependent variable. To encompass different models , we have considered both discriminative and generative models.Following are the keys points where discriminative and generative models differ :-

\begin{itemize}
\item Fitting Technique -  Generally generative models require less sophisticated technique to fit the model i.e. naive bayes and LDA but discriminative models require more complex techniques such as convex and non convex optimization techniques for Lasso , Ridge , Logistic Regression and Sparse Net.

\item Class based Training - Discriminative models require retraining of the complete model again while for generative model separate training is required for each class.

\item Missing Value Treatment -  Missing Values treatment is more difficult for discriminative model as we estimate parameters given x but generative models have simple methods to deal with this problem. 
\item Backward Computation - For generative model , we can infer about the inputs given y as we are making assumptions over the distribution of x and this is not possible for discriminative models.

\item Training Data Requirement - If assumptions are correct generative models require less training data to attain similar performance compared to discriminative models but when assumptions are wrong discriminative models can provide better results compared to generative models.

\item Feature Engineering - As we are making assumption over the input features , any meaningful transformation of x distribution is difficult as it becomes correlated or  violate the assumptions, however its feasible for discriminative models.

\item Probabilities Calibration - Generally generative models produce extreme probabilities because of assumptions however discriminative models produce better calibrated probabilities estimates. 
\end{itemize}

We have considered following discriminative and generative models:-

\begin{tabular}{ |p{5cm}||p{5cm}|}
 \hline
 \multicolumn{2}{|c|}{Supervised Model} \\
 \hline
 Discriminative Model& Generative Model\\
 \hline
Xgboost & Naive Bayes\\
Support Vector Machines & Linear Discriminate Analysis\\
K-Nearest-Neighbor & Quadratic Discriminate Analysis\\
Logistic Elastic Net Classifier & \\
LightGBM & \\
Random Forest & \\
 \hline
\end{tabular}

We have described the above models briefly below:-

\textbf{Extreme Gradient Boosting}:- \\
It is a tree ensemble model where we regularize over leaves of the tree and  each tree created by fitting residual calculated after adding the previous trees. Therefore, it outputs weighted sum of the predictions of multiple regression/classification trees. Relevant notation - $\hat{y_i}$ is the prediction from our Model for ith observation.$\phi(x_i)$ is the prediction function and f represents the classification tree.$\Omega(f)$ is the regularization over the leaves of the trees. Followings are the formula:-

For estimation function - 

\begin{equation}
\hat{y_i} = \phi(x_i) = \sum_{k=1}^K f_k(x_i) 
\end{equation}

Loss Function - 
\begin{equation}
L(\phi) = \sum_{i}l(\hat{y_i},y_i) + \sum_k \Omega(f_k)
\end{equation}

where,
\begin{equation}
\Omega(f) = \gamma T + \frac{1}{2}\lambda\vert\vert w \vert\vert
\end{equation}
\vspace{10 mm}

\textbf{Support Vector Machines}:- \\

It is special machine learning model where we find the hyper plane which gives maximum distance between the classes. Training points which lead to creation of hyper-plane are called support vectors. Assuming we have have two independent variables, then we can represent the hyperplane using following equation:-

\begin{equation}
y = w_0 + w_1x_1 + w_2x_2
\end{equation}

By applying optimization over maximum margin hyperplane:-

\begin{equation}
y = w_0+ \sum_{i}\alpha_iy_ix_i.x 
\end{equation}

For non-separable classes , we can use kernel trick where it can be represented using following equation:-

\begin{equation}
y = w_0 + \sum_{i}\alpha_iy_iK(x_i,x)
\end{equation}

Above optimization is equivalent to solving linear constrained quadratic programming with lower bound of 0 for $\alpha_i$. 
\vspace{10 mm}

\textbf{K Nearest Neighbor} \\

K Nearest Neighbor Classifier is considered lazy learning method. We don't need to fit the model. Given $x_1$ , we find the k-nearest points to $x_1$ and classify based on majority votes. We estimate K based on cross-validation technique leading to best performance. There are different kind of distances that can be considered like manhattan distances and euclidean distances. Euclidean distance has been used. 
\vspace{10 mm}

\textbf{Logistic Elastic Net Classifier} \\

It is regularized classifier with L1 and L2 penalty. It is useful to do automatic variable selection. L1 penalty works best in the sparse solution of variables and L2 works best with multi-collinear variables.

\begin{equation}
\hat{\beta} = \sum_{i=1}^n(-y_i\beta^Tx_i + ln(1+e^{y_i\beta^Tx_i})) + \alpha\lambda\vert\vert\beta\vert\vert +  (1-\alpha)\lambda\vert\vert\beta\vert\vert_2^2
\end{equation}
\vspace{10 mm}

\textbf{LightGBM}
\\
LightGBM is similar to other tree based models.It uses histogram-based algorithms which bucket continuous feature (attribute) values into discrete bins. This speeds up training and reduces memory usage. It allows leaf-wise growth compared to level-wise growth. It deals with categorical features differently with sorting the categories based on $sum \ gradient/sum \ hessian$ and then finding best split based on sorted histogram. 
\\
\vspace{10 mm}
\textbf{Random Forest}
\\
Random Forests are an ensemble learning method for classification. It avoids over fitting by fitting multiple trees. It reduces the variance using bagging idea but increases the bias at the same time. The training algorithm applies the general technique of bootstrap aggregating. It selects randomly B samples with replacement and then fit the deep tree and after that prediction made on the unseen data is the average of B trees.
\\
\vspace{10 mm}
\textbf{Naive Bayes}
\\
Naive bayes is a classification method based on bayes theorem where each feature is independent among themselves given y. It is easy to fit this classification method as we don't have to optimize the parameters. Assuming y being the class variable and $x_1,...,x_n$ are feature vectors.

\begin{equation}
    P(y|x_1,...,x_n) = \frac{P(y)P(x_1,...,x_n|y)}{P(x_1,...,x_n)}
\end{equation}

Assuming independent features,

\begin{equation}
    P(y|x_1,...,x_n) = \frac{P(y)\prod\limits_{i = 1}^{n} p(x_i|y)}{P(x_1,...,x_n)}
\end{equation}

\begin{equation}
    P(y|x_1,...,x_n) 	\propto P(y)\prod\limits_{i = 1}^{n}                                p(x_i|y)
\end{equation}

Above equation is equivalent to below equation

\begin{equation}
    \hat{y}  \propto P(y)\prod\limits_{i = 1}^{n}                                p(x_i|y)
\end{equation}
\\
\vspace{10 mm}
\textbf{Linear Discriminate Analysis}
\\
Linear Discriminate Analysis is a generative classification method. It makes an assumption of normality and independence among features. With strong assumption of normality, it doesn't support categorical features. We separately fit the distribution for each class.

Consider $\vec{x}$ are the features for each sample with y being the response. We model $p(\vec{x}|y=0)$ and $p(\vec{x}|y=1)$ model separately, following normal distribution with $(\vec{\mu_0}, \Sigma_0)$ and $(\vec{\mu_1}, \Sigma_1)$ respectively and based on bayesian optimal threshold , we calculate the boundary of the classifier. Given observation belongs to second class based on the formula below:-

\begin{equation}
({\vec{x}}-{\vec{\mu}}_{0})^{T}\Sigma_{0}^{-1}({\vec {x}}-{\vec {\mu }}_{0}) + \ln \vert\Sigma _{0}\vert-({\vec {x}}-{\vec {\mu }}_{1})^{T}\Sigma_{1}^{-1}({\vec {x}}-{\vec {\mu }}_{1})-\ln \vert \Sigma _{1}\vert > T
\end{equation}

\textbf{Quadratic Discriminate Analysis}
\\
This method holds similar assumptions as Linear Discriminate Analysis, with only exception of different variance. This can also accommodate interaction features.

\subsection{\large{Cross-Validation}}

Cross-Validation is used to estimate the robust hyper-parameters , prediction on test data and calculating generalized error of the algorithm. There are different ways of doing cross-validation such as K-Fold , Stratified Cross Validation ,Leave-One-Out and many more. Good Cross-Validation is the one which doesn't over-fit and performs better on production time-period. But generally we tend to overfit in Finance through Cross-Validation. We split the data into training set and Validation set, each observation belongs to either group to prevent data leak. 

It has been shown that K-Fold CV provides lower cross-validation loss but it might lead to worse performance during production trading or during live trading sessions. There are few reasons for that , one of the leading reason is observations are assumed to be following IID property. When we are doing K-Fold CV , testing dataset gets used multiple times leading to selection and bias. 

One of the K-Fold CV method is  time-series nested  cross-validation technique.

Figure~\ref{fig:foo}
\begin{figure}[!h]
\includegraphics[scale=0.4]{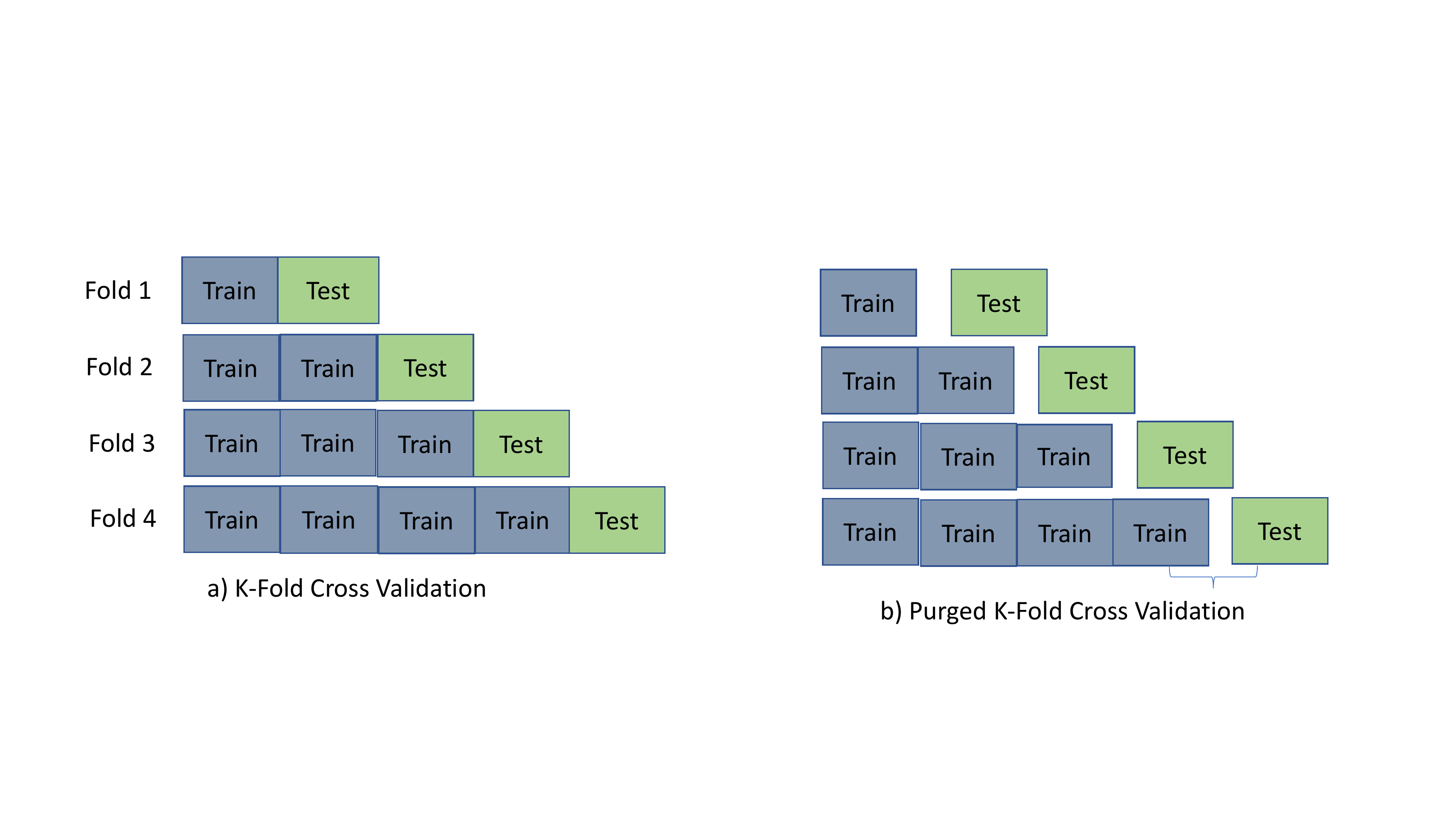}
\caption{Cross-Validation Type}
\label{fig:foo}
\end{figure}

First part shows the vanilla K-Fold Cross Validation method. Second one is Purged Cross-Validation method where we delete some part of the interfering dataset between one Train and Test. This will reduce the leakage model development.

Following is the Fold-wise time-period distribution:-

\begin{center}
\begin{tabular}{ |p{5cm}|p{5cm}|p{5cm}|}
 \hline
 \multicolumn{3}{|c|}{\textbf{Fold Distribution}} \\
 \hline
 \textbf{Fold No} & \textbf{Training Period} & \textbf{Test Period}\\
 \hline
Fold 1 & Aug-Oct 2017 & Nov-2017   \\
Fold 2 & Aug-Nov 2017 & Dec-2017   \\
Fold 3 & Aug-Dec 2017 & Jan-2018   \\
Fold 4 & Aug-2017 - Jan-2018 & Feb-2018 \\
Fold 5 & Aug-2017 - Feb-2018 & Mar-2018   \\
 \hline
\end{tabular}
\end{center}

We have deleted one week data between Training Period and Test Period to incorporate Purging.

\subsection{\large{Stacking - Using One Hidden Layer}}

[6] talks about technique to reduce generalization error rate. It aims to achieve generalization accuracy by combining weak learner. It is considered to be more sophisticated than winner-takes-all strategy. Lets assume {$T_i$} be one of the generalizer then we combine by taking the output of each generalizer and making it a new space. We can learn the estimates by in-sample/out-sample techniques. It can also be considered a flexible version of cross-validation.

It generally creates different levels of models having one level of output being the input for next level. Primarily , it removes the biases of all models leading to generalization of all the models. Many also consider it as "black art" as we have multiple options of keeping diverged models at each level but it has been very effective in producing stable and effective models.

Similarly, stacked regression[7] was introduced where we linearly  combine different predictors to improve accuracy. It also introduces non-negativity constraints over the coefficients. Assuming we have K predictors or generalizers with (y,x) being the given data. $T_1(x),T_2(x),...,T_K(x)$ are the K generalizers and we combine each generalizer to create level-1 data. We leave one fold data to create level-1 data, getting $u_k^{(-t)}(x) ,k = 1,...,K$ , leading to K-variable input space.

\begin{equation}
z_kt =  u_k^{(-t)}(x_t)
\end{equation}

Finally, Level-one data is created as $(y_t,z_t) , t = 1,...,n$, [7] propose to linear combination of each generalizer and parameters are estimated using least square error. It addresses two problems of over-fitting and multi-collinearity by using level-one data and adding non-negative constraints over coefficients.

\begin{equation}
u(x) = \sum_k\beta_ku_k(x)
\end{equation}

We have similar motivation but we have used non-linear combination of generalizer by inducing hidden layer. It has been shown that hidden layer can be very effective in learning non-linear functions.

Level 0 has 7 models and Level 1 is hidden layer with 6 nodes. We first trained level 0 models from Aug-2017 to Mar-2018. We trained complete model with Level 1 hidden layer from Apr-2018 to June-2018. We have predicted the output as 0/1 for each model in level 0 and final loss is logloss.

Following is the flow for stacked generalizer:-

\begin{figure}[!h]
\includegraphics[scale=0.4]{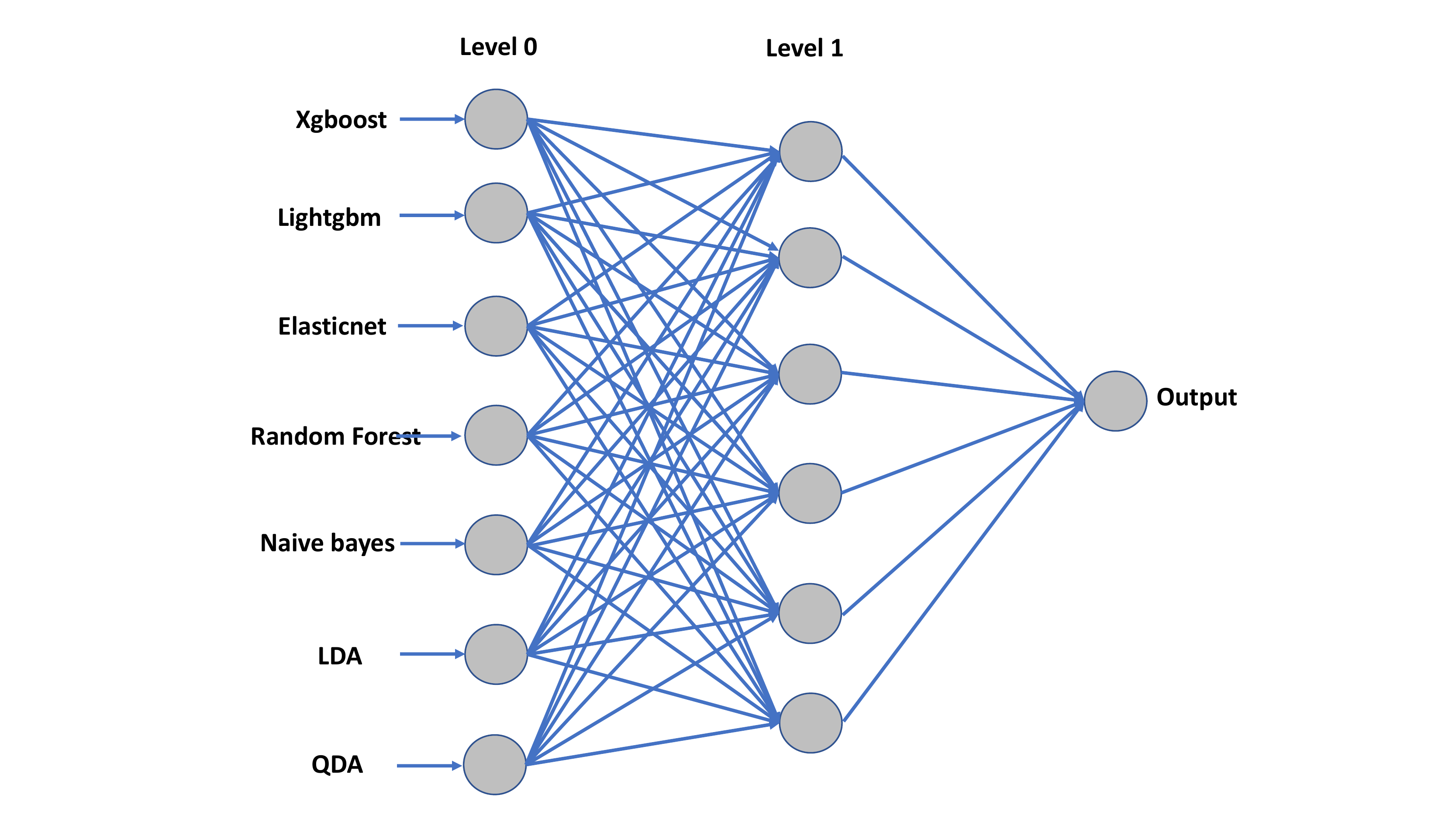}
\caption{Stacking with Neural Network}
\label{fig:foo}
\end{figure}

\subsection{\large{Feature Selection}}

For tree models Feature importance can be used as Feature selection methodology. Following are the ways to calculate feature importance:-

\begin{itemize}
    \item Gain - In spliting features , we calculate the decrease in gini impurity or entropy which is finally combined to create combined Gain for each feature. 
    \item Real Cover - Similarly , when features are splitted, the split occurs over observations. We count the observations where split occured and it is finally combined leading to Real Cover.
\end{itemize}

[8] and [9] talks about visualizing the feature importance by estimating the variability in the estimated function by varying each particular variable and keeping other variables at their average. It is called partial dependence plot. 

Partial dependence function for regression is defined below:-

\begin{equation}
    \hat{f}_{x_t}(x_t) = E_{x_{-t}}[\hat{f}(x_t,x_{-t})] = \int \hat{f}({x_t},x_{-t})dP x_{-t}
\end{equation}

Since calculating the integration is difficult, we use Monte Carlo to estimate the partial function.

\begin{equation}
    \hat{f}_{x_t}(x_t) = \frac{\sum_{i=1}^n \hat{f}_{x_t}(x_t,x_{-t}^{(i)})}{n}
\end{equation}

It comes with rigid assumption that feature t is not correlated with rest of the features. For classification , we output probabilities rather than 0/1.

Recently,[10] talks about partial dependence-based variable importance measure. It calculates the standard deviation for continuous predictors and range statistics divided by 4. Lets assume importance measure is i for any variable then 

\[
 i(x_1) = 
  \begin{cases} 
   \sqrt{\frac{\sum_{i=1}^n[\hat{f}(x_1i)-\bar{\hat{f}}(x_{1i})]^2}{n-1}} & \text{if } x_1 \ is \ continuous \\
   \frac{max_{i}(\hat{f}(x_{1i})) - min_{i}(\hat{f}(x_{1i}))}{4} & \text{if } x_1 \ is \ categorical
  \end{cases}
\]

For Stacked generalization, we first analyzed level-1 to calculate variable importance for each generalizer and in second step , important generalizer is analyzed further to calculate the variable importance and finally variable importance is calculated by taking average of all.
We have K-generalizer then $i(x_t)$ is calculated as below:-

\begin{itemize}
    \item $i(T_k)$ is calculated based on above formula
    \item $w_k = \frac{i(T_k)}{\sum_{k=1}^K i(T_k)}$
    \item $i_k(x_t)$ is calculated based on above formula for each ${T_k}$
    \item $i(x_t) = \sum_{k=1}^K w_k*i_k(x_t)$
\end{itemize}

This is model independent variable importance for each feature. Its important to consider weight here as it is 2 level model.

\section{\Large{Results}}
\setcounter{section}{3}
\subsection{Performance}

\begin{flushleft}
\textbf{Extreme Gradient Boosting(Xgboost)} requires $max \ depth$, $min \ child \ weight$, $subsample$, $colsample \ bytree$, $reg \ alpha$, $reg \ lambda$, $n \ estimators$ and $learning \ rate$ to be estimated. There are many methods to tune hyper-parameters mentioned above. Grid-Search cross-validation , random search and bayesian optimization are among those. Generally Grid-Search cross-validation and bayesian optimization requires lot of time tune parameters. Therefore, we have used random search method to tune parameter. We have used 100 of iteration across each fold to estimate the best hyper-parameter. Cross-validation results are provided in the supplementary materials.
In previous we have mentioned performance periods where first level-0 data was created and then the output is used for level-1 and final performance of stacking is compared with the rest of the models. 

Apr-May 2018 time period has been used to create level-1 data in stacking. Stacked generalizer's performance compared to rest of the models during June - July 2018. Below is the performance for \textbf{xgboost}:-

\centering 
\begin{tabular}{|p{5cm}|p{5cm}|p{5cm}|}
 \hline
 \multicolumn{3}{|c|}{\textbf{Xgboost}} \\
 \hline
 Parameter & Apr-May 2018 & June - July 2018\\
 \hline
 
 AUC       & 0.59 & 0.46 \\
 Accuracy  & 0.57 & 0.46 \\
 Precision & 0.59 & 0.48 \\
 Recall    & 0.59 & 0.59 \\
 F1        & 0.59 &  0.53 \\
 \hline
\end{tabular}

\textbf{Support Vector Machines(SVM)} requires different kernel, gamma and cost as hyper-parameters to be tuned. For non-separating hyper-planes, kernel trick is important where we use  radial basis function and sigmoid to map into different space to make it separable. Cross-validation results are provided in the supplementary materials. Below is the performance for \textbf{SVM}:-

\centering 
\begin{tabular}{|p{5cm}|p{5cm}|p{5cm}|}
 \hline
 \multicolumn{3}{|c|}{\textbf{SVM}} \\
 \hline
 Parameter & Apr-May 2018 & June - July 2018\\
 \hline
 AUC       & 0.39 & 0.53 \\
 Accuracy  & 0.48 & 0.50 \\
 Precision & 0.51 & 0.53 \\
 Recall    & 0.62 & 0.59 \\
 F1        & 0.56 &  0.56 \\
 \hline
\end{tabular}

\textbf{K-Nearest-Neighbor(KNN)} requires number of neighbors to be tuned. Cross-validation results are provided in the supplementary materials. Below is the performance for \textbf{KNN}:-

\begin{tabular}{|p{5cm}|p{5cm}|p{5cm}|}
 \hline
 \multicolumn{3}{|c|}{\textbf{KNN}} \\
 \hline
 Parameter & Apr-May 2018 & June - July 2018\\
 \hline
 AUC       & 0.53 & 0.52 \\
 Accuracy  & 0.59 & 0.52 \\
 Precision & 0.59 & 0.52 \\
 Recall    & 0.66 & 0.52 \\
 F1        & 0.62 & 0.52 \\
 \hline
\end{tabular}

\textbf{Light Gradient Boosting(LightGBM)} requires $learning \ rate$,$subsample \ gen$, $subsample \ freq$, $colsample \ bytree$, $reg \ alpha$, $reg \ lambda$, $max \ depth$, $min \ child \ weight$, $num \ leaves$ and $n \ estimators$. Cross-validation results are provided in the supplementary materials. Below is the performance for \textbf{LightGBM}:-

\begin{tabular}{|p{5cm}|p{5cm}|p{5cm}|}
 \hline
 \multicolumn{3}{|c|}{\textbf{LightGBM}} \\
 \hline
 Parameter & Apr-May 2018 & June - July 2018\\
 \hline
 AUC       & 0.62 & 0.52 \\
 Accuracy  & 0.61 & 0.52 \\
 Precision & 0.63 & 0.52 \\
 Recall    & 0.69 & 0.52 \\
 F1        & 0.66 & 0.52 \\
 \hline
\end{tabular}

\textbf{Random Forest(RF)} requires $n \ estimators$,$max \ depth$,$min \ samples \ split$ and $min \ samples \ leaf$. Cross-validation results are provided in the supplementary materials. Below is the performance for \textbf{RF}:-

\begin{tabular}{|p{5cm}|p{5cm}|p{5cm}|}
 \hline
 \multicolumn{3}{|c|}{\textbf{RF}} \\
 \hline
 Parameter & Apr-May 2018 & June - July 2018\\
 \hline
 AUC       &  0.55  & 0.50 \\
 Accuracy  &  0.55  & 0.50 \\
 Precision &  0.55  & 0.50 \\
 Recall    &  0.76  & 0.59 \\
 F1        & 0.64   & 0.54 \\
 \hline
\end{tabular}

\textbf{Logistic Elastic Net(LogisticENet)} requires $alpha$ and $l1 \ ratio$. Cross-validation results are provided in the supplementary materials. Below is the performance for \textbf{LogisticENet}:-

\begin{tabular}{|p{5cm}|p{5cm}|p{5cm}|}
 \hline
 \multicolumn{3}{|c|}{\textbf{LogisticENet}} \\
 \hline
 Parameter & Apr-May 2018 & June - July 2018\\
 \hline
 AUC       &  0.53  & 0.50 \\
 Accuracy  &  0.52  & 0.50 \\
 Precision &  0.52  & 0.48 \\
 Recall    &  0.48  & 0.44 \\
 F1        &  0.50  & 0.46 \\
 \hline
\end{tabular}

\textbf{Naive Bayes(NB)} doesn't require any hyper-parameters to be tuned. Below is the performance for \textbf{NB}:-

\begin{tabular}{|p{5cm}|p{5cm}|p{5cm}|}
 \hline
 \multicolumn{3}{|c|}{\textbf{NB}} \\
 \hline
 Parameter & Apr-May 2018 & June - July 2018\\
 \hline
 AUC       &  0.73  & 0.41 \\
 Accuracy  &  0.52  & 0.50 \\
 Precision &  0.68  & 0.40 \\
 Recall    &  0.66  & 0.44 \\
 F1        &  0.67  & 0.42 \\
 \hline
\end{tabular}

\textbf{Linear Discriminate Analysis(LDA)} doesn't require any hyper-parameters to be tuned. Below is the performance for \textbf{LDA}:-

\begin{tabular}{|p{5cm}|p{5cm}|p{5cm}|}
 \hline
 \multicolumn{3}{|c|}{\textbf{LDA}} \\
 \hline
 Parameter & Apr-May 2018 & June - July 2018\\
 \hline
 AUC       &  0.62  & 0.48 \\
 Accuracy  &  0.54  & 0.48 \\
 Precision &  0.61  & 0.50 \\
 Recall    &  0.66  & 0.48 \\
 F1        &  0.63  & 0.49 \\
 \hline
\end{tabular}

\textbf{Quadratic Discriminate Analysis(QDA)} doesn't require any hyper-parameters to be tuned. Below is the performance for \textbf{QDA}:-

\begin{tabular}{|p{5cm}|p{5cm}|p{5cm}|}
 \hline
 \multicolumn{3}{|c|}{\textbf{QDA}} \\
 \hline
 Parameter & Apr-May 2018 & June - July 2018\\
 \hline
 AUC       &  0.59  & 0.55 \\
 Accuracy  &  0.55  & 0.52 \\
 Precision &  0.60  & 0.56 \\
 Recall    &  0.52  & 0.81 \\
 F1        &  0.56  & 0.67 \\
 \hline
\end{tabular}

Performance of generative and discriminative models are very similar with best performance from Quadratic Discriminative Analysis(QDA)

\textbf{Stacked Generalization(SG)} requires number of layers and number of nodes in the hidden layer to be tuned. Below is the performance for \textbf{SG}:-

\begin{tabular}{|p{5cm}|p{5cm}|p{5cm}|}
 \hline
 \multicolumn{3}{|c|}{\textbf{SG}} \\
 \hline
 Parameter & Apr-May 2018 & June - July 2018\\
 \hline
 AUC       &  0.61  & 0.50 \\
 Accuracy  &  0.52  & 0.54 \\
 Precision &  0.61  & 0.52 \\
 Recall    &  0.59  & 0.59 \\
 F1        &  0.60  & 0.55 \\
 \hline
\end{tabular}

We have got best accuracy

\subsection{Feature Selection}

In previous section we showed the methodology to calculate feature importance based on partial dependence plot for stacked generalization. First we have shown the model importance based on stack generalization and then Top 10 individual features importance for each model. We have calculated overall feature importance as defined above.

\begin{figure}[H]
\begin{center}
\includegraphics[scale=0.3]{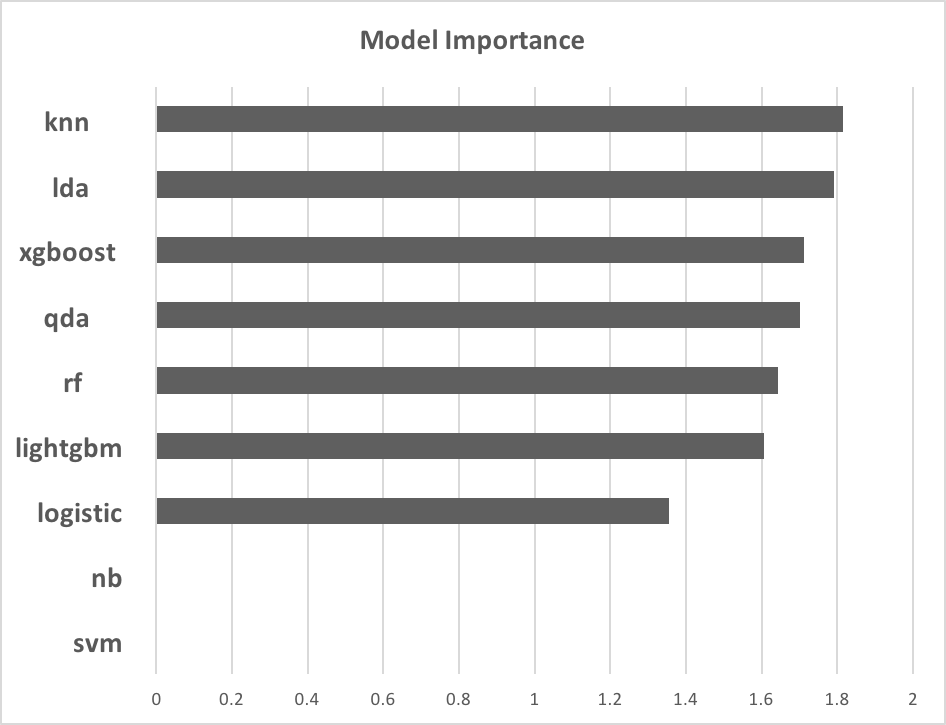}
\caption{Model Importance}
\label{fig:foo}
\end{center}
\end{figure}

\begin{figure}[H]
\begin{subfigure}{0.3\textwidth}
\includegraphics[width=\linewidth]{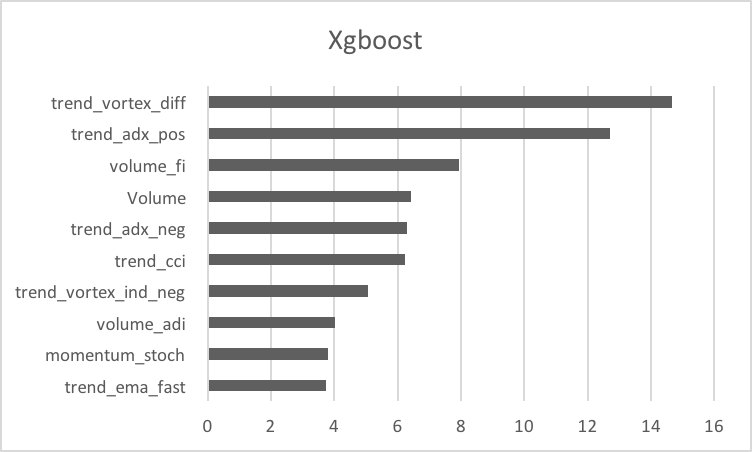}
\end{subfigure}
\begin{subfigure}{0.3\textwidth}
\includegraphics[width=\linewidth]{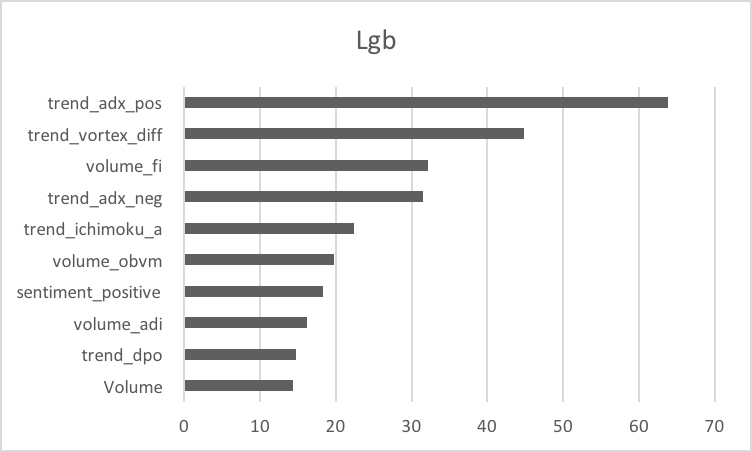}
\end{subfigure}
\begin{subfigure}{0.3\textwidth}
\includegraphics[width=\linewidth]{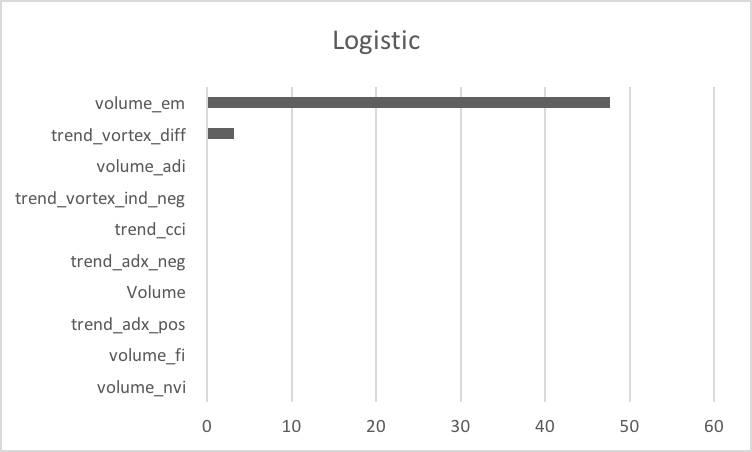}
\end{subfigure}
\caption{Feature importance. Left: Xgboost. Middle: LightGBM. Right: Logistic Regression.}
\label{fig:foo}
\end{figure}

\begin{figure}[!ht]
\begin{subfigure}{0.3\textwidth}
\includegraphics[width=\linewidth]{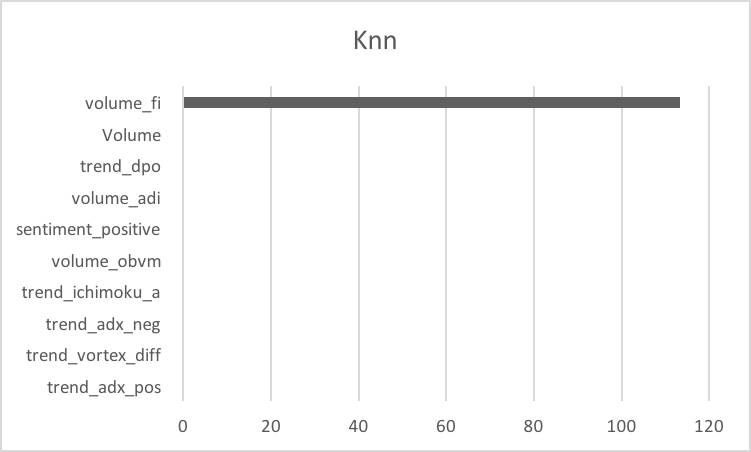}
\end{subfigure}
\begin{subfigure}{0.3\textwidth}
\includegraphics[width=\linewidth]{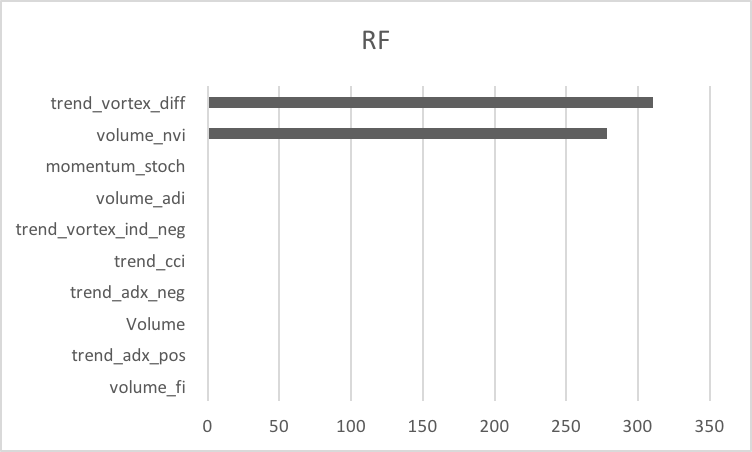}
\end{subfigure}
\begin{subfigure}{0.3\textwidth}
\includegraphics[width=\linewidth]{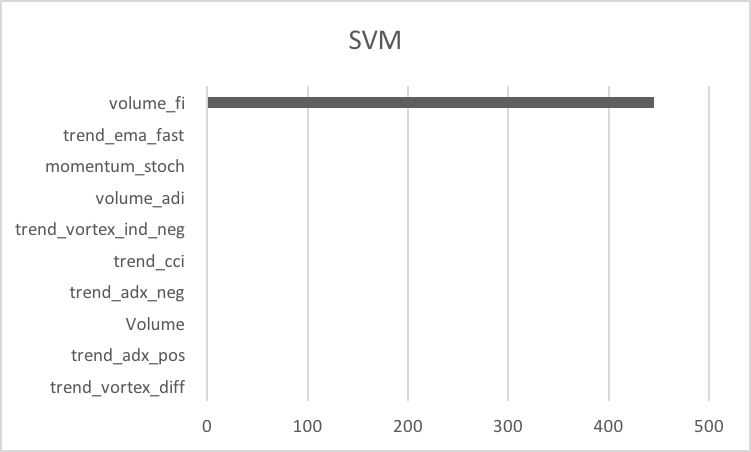}
\end{subfigure}
\caption{Feature importance. Left: KNN. Middle: RF. Right: SVM.}
\label{fig:foo}
\end{figure}

\begin{figure}[!ht]
\begin{subfigure}{0.3\textwidth}
\includegraphics[width=\linewidth]{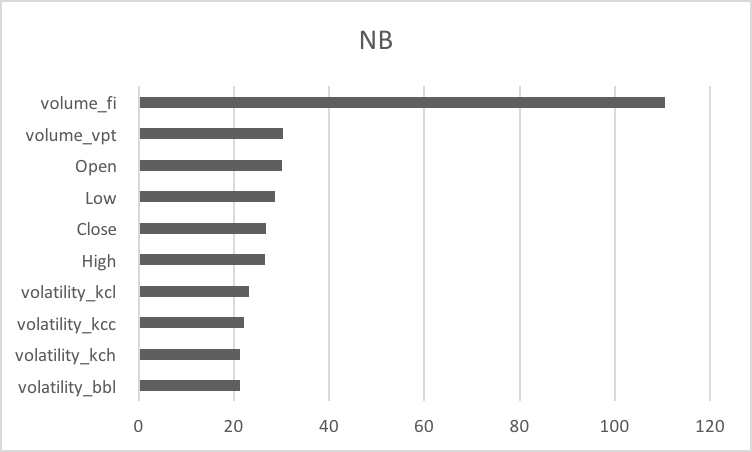}
\end{subfigure}
\begin{subfigure}{0.3\textwidth}
\includegraphics[width=\linewidth]{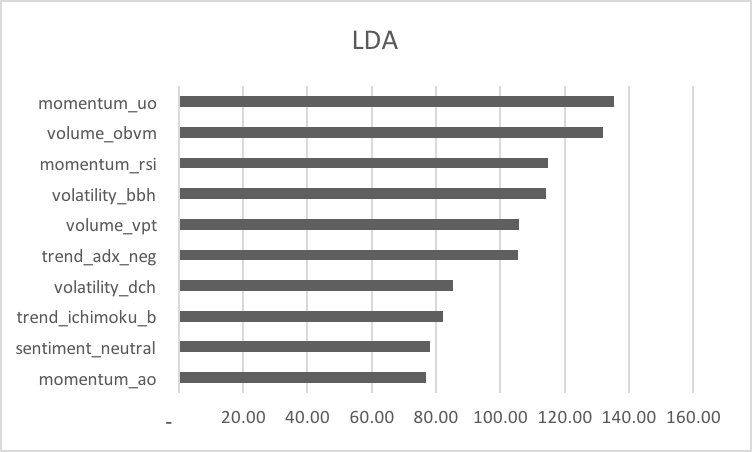}
\end{subfigure}
\begin{subfigure}{0.3\textwidth}
\includegraphics[width=\linewidth]{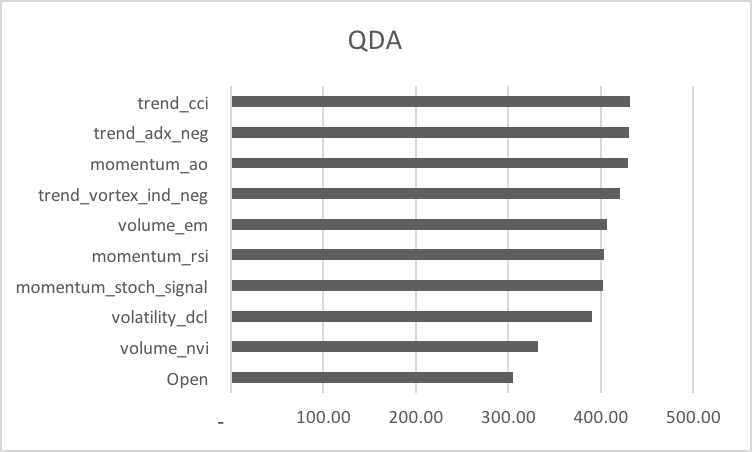}
\end{subfigure}
\caption{Feature importance. Left: NB. Middle: LDA. Right: QDA.}
\label{fig:foo}
\end{figure}

\begin{figure}[!ht]
\begin{center}
\includegraphics[scale=0.3]{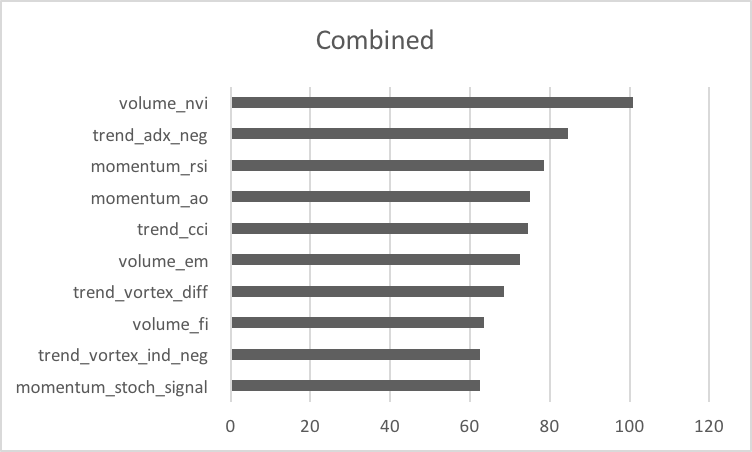}
\caption{Overall Feature Importance}
\label{fig:foo}
\end{center}
\end{figure}

Naive bayes(NB) and Support Vector Machines(SVM) are not contributing to Stacked model with highest contribution from K nearest neighbor(KNN) and linear discriminate analysis(LDA).

For Xgboost, most important features are $trend$ features. LightGBM has $sentiment \ positive$ as important features also. Logistic ElasticNet has only two important features $volume em$ and $trend \ vortex \ diff$. KNN has only important feature $volume \ fi$. Similarly for Random Forest(RF) , $trend \ vortex \ diff$ and $volume nvi$. Support vector Machines (SVM) has $volume \ fi$ as important features. Naive Bayes(NB) has mostly volatility features as important features. Linear Discriminate Analysis(LDA) and Quadratic Discriminate Analysis(QDA) have important features distributed across momentum,trend,volume,volatility and sentiment.

\end{flushleft}

\section{Conclusions}
\begin{flushleft}
Cryptocurrencies direction prediction can be improved by using different generative and discriminative models.The challenge has been to identify the domain of features and producing generalized model to perform well across the time-period with different natures of it. Cross-Validation is very important for building robust model and its more tricky with time-series data set. Purged Cross-Validation addresses these problems. Stacked generalization technique has been used to create generalized model which has more information compared to individual models leading to better accuracy. Interpreting the machine learning models have become utmost important exercise. Partial dependence plots(PDPs) is used to uncover each model important features and also contribution from each model to stacked model. Having multiple models, we have used new definition based on PDPs to create combined feature importance. Finally, important features can be used to do day trading as well.   
\end{flushleft}

\section{Appendix}

\subsection{Volume Technical Indicators}

\textbf{Accumulation/Distribution Index (ADI)} -
It is a combination of price and volume and leading indicator. \\
\begin{equation}
CLV_t = \frac{(close_t - low_t) - (high_t - close_t)}{(high_t - low_t)} 
\end{equation}
\begin{equation}
accdist_t = accdist_{t-1} + volume_t\times CLV_t
\end{equation}

\textbf{On balance volume} - It is based on total cumulative volume.

\begin{equation}
OBV_t = OBV_{t-1} + \begin{cases}
                    Volume_t, & if close_t > close_{t-1} \\
                    0 & close_t = close_{t-1} \\
                    -Volume_t & close_t < close_{t-1}
                   \end{cases}
\end{equation}

\textbf{On balance volume mean} - It is 10 days rolling mean of On balance volume indicator.

\textbf{Chaikin money flow} -  It measures the amount of Money Flow Volume over a specific period. We have considered sum of 20 days for the indicator. 

\begin{equation}
MFM_t = \frac{(close_t - low_t) - (high_t - close_t)}{(high_t - low_t)}
\end{equation}

\begin{equation}
MFV_t = MFM_t*volume_t 
\end{equation}

\begin{equation}
CMF_t = \frac{EMA(MFV_t,20)}{EMA(volume_t,20)}
\end{equation}

\textbf{Force index} - It shows the buying and selling pressure present.

\begin{equation}
FI    =  {Close -  Close_{prev-1}}\times Volume - Volume_{prev-1}
\end{equation}

\textbf{Ease of movement} - It relates an asset's price change to its volume

\begin{equation}
EMV = ((high - high_{prev}) + (low - low_{prev})) * (high - low) / (2 * volume)
\end{equation}
\begin{equation}
EMV_t = 20 \ Rolling \ Moving \ Average \ of EMV
\end{equation}

\textbf{Volume Price Trend} -  It is based on running cumulative volume that adds or substracts a mutiple of change in close price.

\begin{equation}
VPT = VPT_{prev} + volume\times \frac{close - close_{prev}}{close_{prev}}
\end{equation}

\textbf{Negative volume index} - 
It is about detecting smart money being active using volume.

\begin{equation}
nvi_t = 
\begin{cases}
nvi_{t-1} \times ( 1 + \frac{(close_t - close_{t-1})}{close_{t-1}}, &  volume_t > volume_{t-1} \\
nvi_{t-1} & volume_t <= volume_{t-1}
\end{cases}
\end{equation}

\subsection{Volatility Technical Indicators}

\textbf{Average true range} - The indicator provide an indication of the degree of price volatility.

\begin{equation}
ATR_t = max(high_t,close_{t-1}) - max(low_t,close_{t-1})
\end{equation}
\begin{equation}
ATR_t = EMA(ATR_t,20)
\end{equation}

\textbf{Bollinger Moving Average} - It is the moving average of close price.

\begin{equation}
BMVG_t = MA(close_t,20)
\end{equation}

\textbf{Bollinger Lower Band} - It is lower band at 2 times an 20-period standard deviation below the moving average of 20 days.

\begin{equation}
BMVG_t  = MA(close_t,20)
\end{equation}
\begin{equation}
sigma_t = \sigma(close_t,20)
\end{equation}
\begin{equation}
BBLband_t = BMVG_t - 2 * sigma_t
\end{equation}

\textbf{Bollinger Higher Band} - It is higher band at 2 times an 20-period standard deviation above the 20 days moving average of close price.

\begin{equation}
BMVG_t  = MA(close_t,20)
\end{equation}
\begin{equation}
sigma_t = \sigma(close_t,20)
\end{equation}
\begin{equation}
BBLband_t = BMVG_t + 2 * sigma_t
\end{equation}

\textbf{Bollinger Higher Band Indicator} - It returns 1, if close is higher than bollinger high band. Else, return 0

\textbf{Bollinger Lower Band Indicator} - It returns 1, if close is lower than bollinger lower band. Else, return 0

\textbf{Keltner Channel Central} - It is  10-day simple moving average of typical price.

\begin{equation}
tp_t  = \frac{(high + low + close)}{3.0} 
\end{equation}
\begin{equation}
kcc_t = MA(tp_t,20)
\end{equation}

\textbf{Keltner Channel Higher Band} - It shows a simple moving average line (high) of typical price.

\begin{equation}
tp_t = \frac{((4 * high) - (2 * low) + close)}{3.0}
\end{equation}
\begin{equation}
kch_t = MA(tp_t,10)   
\end{equation}

\textbf{Keltner Channel Lower Band} - It shows a simple moving average line (low) of typical price.

\begin{equation}
tp_t  = \frac{((-2 * high) + (4 * low) + close)}{3.0} 
\end{equation}
\begin{equation}
kcl_t = MA(tp_t,10)    
\end{equation}

\textbf{Keltner Channel Higher Band Indicator} - It return 1 if close price is greater than $kch_t$,else 0.

\textbf{Keltner Channel Lower Band Indicator} - It return 1 if close price is lower than $kcl_t$,else 0.

\textbf{Donchian Channel Higher Band} - The upper band shows the highest price of an asset for 20 periods.

\begin{equation}
dch_t = max(close_t,close_{t-1},...close_{t-19})
\end{equation}

\textbf{Donchian Channel Lower Band} - The lower band shows the highest price of an asset for 20 periods.

\begin{equation}
dcl_t = min(close_t,close_{t-1},...close_{t-19})
\end{equation}

\textbf{Donchian Channel Higher Band Indicator} - It returns 1 if close is greater than $dch_t$. 

\textbf{Donchian Channel Lower Band Indicator} - It returns 1 if close is lower than $dcl_t$

\subsection{Trend}

\textbf{Moving Average Convergence Divergence(MACD)}- It is  a trend-following momentum indicator that shows the relationship between fast and slow moving averages of prices.

\begin{equation}
Emafast_t = EMA(close_t,12)
\end{equation}
\begin{equation}
Emaslow_t = EMA(close_t,26)
\end{equation}
\begin{equation}
MACD_t    = Emafast_t - Emaslow_t    
\end{equation}

\textbf{Moving Average Convergence Divergence Signal} -  It is EMA of MACD

\textbf{Moving Average Convergence Divergence Diff} - It is difference between MACD and MACD Signal

\textbf{Exponential Moving Average} - It is exponential moving average of close price

\textbf{Average Directional Movement Index(ADX)} - It is 14 day averages of the difference between +DI and -DI, and indicates the strength of the trend.

\begin{equation}
tr_t = max(high_t ,close_{t-1}) - min(low_t,close_{t-1})
\end{equation}

\begin{equation}
trs_t = Sum(tr,20)
\end{equation}

\begin{equation}
up_t = high_t - high_{t-1}
\end{equation}

\begin{equation}
dn_t = low_{t-1} - low_t
\end{equation}

\begin{equation}
pos_t = ((up_t > dn_t) and (up_t > 0)) \times up_t
\end{equation}

\begin{equation}
neg_t = ((dn_t > up_t) and (dn_t > 0)) \times dn_t
\end{equation}

\begin{equation}
dip_t = 100 \times \frac{Sum(pos_t,n)}{trs_t}
\end{equation}

\begin{equation}
din_t = 100 \times \frac{Sum(neg_t,n)}{trs_t}
\end{equation}

\begin{equation}
dx_t = 100 \times abs(\frac{(dip_t - din_t)}{(dip_t + din_t)})
\end{equation}

\begin{equation}
adx_t = EMA(dx_t,14)
\end{equation}

\textbf{Average Directional Movement Index Positive (ADX)} - It is +DI($dip_t$) \\
\textbf{Average Directional Movement Index Negative (ADX)} - It is -DI($din_t$) \\
\textbf{Average Directional Movement Index Indicator (ADX)} - It returns 1 if difference between +DI and -DI greater than 0 and else 0. 
\\
\textbf{Vortex Indicator Positive (VI)} - It captures bullish signal when positive oscillator trend crosses  negative oscillator trend.
\\
\begin{equation}
tr = max(high_t,close_{t-1}) - min(low_t,close_{t-1})
\end{equation}
\begin{equation}
trn_t = Sum(tr_t,14)
\end{equation}
\begin{equation}
vmp_t = abs(high_t - low_{t-1})
\end{equation}
\begin{equation}
vmn_t = abs(low_t - high_{t-1})
\end{equation}
\begin{equation}
vip_t = \frac{Sum(vmp_t,14)}{trn_t}
\end{equation}

\textbf{Vortex Indicator Negative (VI)} - It captures bearish signal when negative oscillator trend crosses  positive oscillator trend.

\begin{equation}
tr = max(high_t,close_{t-1}) - min(low_t,close_{t-1})
\end{equation}

\begin{equation}
trn_t = Sum(tr_t,14)
\end{equation}
\begin{equation}
vmp_t = abs(high_t - low_{t-1})
\end{equation}
\begin{equation}
vmn_t = abs(low_t - high_{t-1})
\end{equation}
\begin{equation}
vin_t = \frac{Sum(vmn_t,14)}{trn_t}
\end{equation}

\textbf{Trix} - It shows the percent rate of change of a triple exponentially smoothed moving average.

\begin{equation}
EMA1_t = EMA(close_t,14)
\end{equation}
\begin{equation}
EMA2_t = EMA(EMA1_t,14)
\end{equation}
\begin{equation}
EMA3_t = EMA(EMA2_t,14)
\end{equation}
\begin{equation}
trix_t = \frac{(EMA3_t - EMA3_{t-1})}{EMA3_{t-1}}
\end{equation}

\textbf{Mass Index(MI)} - It uses the high-low range to identify trend reversals based on range expansions. It identifies range bulges that can foreshadow a reversal of the current trend.

\begin{equation}
amplitude_t = high_t - low_t
\end{equation}
\begin{equation}
EMA1_t      = EMA(amplitude_t,9)
\end{equation}
\begin{equation}
EMA2_t      = EMA(EMA1_t,26)
\end{equation}
\begin{equation}
mass_t      = \frac{EMA1_t}{EMA2_t}
\end{equation}
\begin{equation}
mass_t      = Sum(mass_t,25)
\end{equation}

\textbf{Commodity Channel Index(CCI)} - CCI measures the difference between a security's price change and its average price change. High positive readings indicate that prices are well above their average, which is a show of strength. Low negative readings indicate that prices are well below their average, which is a show of weakness.

\begin{equation}
pp_t = \frac{(high_t + low_t + close_t)}{3.0} 
\end{equation}

\begin{equation}
cci_t = \frac{(pp_t - MA(pp_t,20))}{0.015*\sigma(pp_t,20)}    
\end{equation}

\textbf{Detrended Price Oscillator (DPO)} - It is an indicator designed to remove trend from price and make it easier to identify cycles.

\begin{equation}
dpo_t = close_{t-10} - MA(close_t,20)
\end{equation}

\textbf{KST Oscillator (KST)} - It is useful to identify major stock market cycle junctures because its formula is weighed to be more greatly influenced by the longer and more dominant time spans, in order to better reflect the primary swings of stock market cycle.
r1=10, r2=15, r3=20, r4=30, n1=10, n2=10, n3=10 and n4=15.
\begin{equation}
rocma_1 = \frac{(close_t - close_{t-r1})}{MA(close_{t-r1},n1)} 
\end{equation}

\begin{equation}
rocma_2 = \frac{(close_t - close_{t-r2})}{MA(close_{t-r2},n2)} 
\end{equation}
\begin{equation}
rocma_3 = \frac{(close_t - close_{t-r3})}{MA(close_{t-r3},n3)} 
\end{equation} 
\begin{equation}
rocma_4 = \frac{(close_t - close_{t-r4})}{MA(close_{t-r4},n4)} 
\end{equation} 
\begin{equation}
kst_t = 100 * (rocma1 + 2 * rocma2 + 3 * rocma3 + 4 * rocma4)
\end{equation}    

\textbf{KST Oscillator (KST Signal)} - It is useful to identify major stock market cycle junctures because its formula is weighed to be more greatly influenced by the longer and more dominant time spans, in order to better reflect the primary swings of stock market cycle

\begin{equation}
kst_{sig_t} = MA(kst_t,9)
\end{equation}

\textbf{Ichimoku Kinkō Hyō A (Ichimoku)} - It identifies the trend and look for potential signals within that trend. $n_1=9,n_2=26$

\begin{equation}
 conv_t = \frac{max(high_t,n_1) + min(low_t,n_1)}{2}
\end{equation} 
\begin{equation}
base_t = \frac{max(high_t,n_2) + min(low_t,n_2)}{2}
\end{equation}\begin{equation}
spana_t = \frac{conv + base}{2}
\end{equation}
\begin{equation}
spana_t = spana_{t-n_2}    
\end{equation}

\textbf{Ichimoku Kinkō Hyō B (Ichimoku)} - It identifies the trend and look for potential signals within that trend.$n_2=26, n_3=52$

\begin{equation}
spanb_t = \frac{max(high_t,n_3) + min(low_t,n_3)}{2}
\end{equation}

\begin{equation}
spana_t = spana_{t-n_2}  
\end{equation}

\subsection{Momentum}
\textbf{Relative Strength Index (RSI)} - It Compares the magnitude of recent gains and losses over a specified time period to measure speed and change of price movements of a security.

\begin{equation}
up_t = 
\begin{cases}
close_t - close_{t-1}, & if close_t > close_{t-1} \\
0 & close_t <= close_{t-1}
\end{cases}
\end{equation}

\begin{equation}
down_t = 
\begin{cases}
close_t - close_{t-1}, & if close_t < close_{t-1} \\
0 & close_t >= close_{t-1}
\end{cases}
\end{equation}

\begin{equation}
rsi_t = 100*\frac{EMA(up_t,14)}{EMA(up_t,14)+EMA(down_t,14)}
\end{equation}

\textbf{True strength index (TSI)} - It Shows both trend direction and overbought/oversold conditions. $r = 25, s = 13$

\begin{equation}  
m   = close_t - close_{t-1}
\end{equation} 
\begin{equation}  
m_1 = EMA(EMA(m,r),s)
\end{equation} 
\begin{equation}  
m_2 = EMA(EMA(\vert m \vert,r),s)
\end{equation} 
\begin{equation}  
tsi = 100*\frac{m_1}{m_2}
\end{equation}    
    
\textbf{Ultimate Oscillator} - A momentum oscillator designed to capture momentum across three different timeframes.
    
\begin{equation} 
bp_t = close_t - min(low_t,close_{t-1})
\end{equation} 

\begin{equation}
tr_t = min(high_t,close_{t-1}) - min(low_t,close_{t-1})
\end{equation}

\begin{equation}
avg\_7_t = \frac{\sum_{i={t-7}}^t bp_i}{\sum_{i={t-7}}^t tr_i}
\end{equation}

\begin{equation}
avg\_14_t = \frac{\sum_{i={t-14}}^t bp_i}{\sum_{i={t-14}}^t tr_i}
\end{equation}

\begin{equation}
avg\_28_t = \frac{\sum_{i={t-28}}^t bp_i}{\sum_{i={t-28}}^t tr_i}
\end{equation}

\begin{equation}
UO_t = 100*\frac{[(4*avg\_7_t)+(2*avg\_14_t)+avg\_28_t]}{(4+2+1)}
\end{equation}

\textbf{Stochastic Oscillator} - Developed in the late 1950s by George Lane. The stochastic oscillator presents the location of the closing price of a stock in relation to the high and low range of the price of a stock over a period of time, typically a 14-day period.
    
\begin{equation}
smin_t = min(low_t,14)
\end{equation}
\begin{equation}
smax_t = max(high_t,14)
\end{equation}
\begin{equation}
stoch_t = 100 * \frac{(close_t - smin_t)}{(smax_t - smin_t)}
\end{equation}
    
\textbf{Stochastic Oscillator Signal} -  It shows SMA of Stochastic Oscillator. Typically a 3 day SMA.

\begin{equation}
stoch_{sig_t} = MA(stoch_t,3)
\end{equation}    
    
\textbf{Williams \%R} - Developed by Larry Williams, Williams \%R is a momentum indicator that is the inverse of the Fast Stochastic Oscillator. Also referred to as \%R, Williams \%R reflects the level of the close 
relative to the highest high for the look-back period. In contrast, the Stochastic Oscillator reflects the level of the close relative to the lowest low. \%R corrects for the inversion by multiplying the raw value by -100. As a result, the Fast Stochastic Oscillator and Williams \%R 
produce the exact same lines, only the scaling is different. Williams \%R oscillates from 0 to -100.
lbp = 14

\begin{equation}
hh_t = max(high_t,lbp)
\end{equation}

\begin{equation}
ll_t = min(low_t,lbp)
\end{equation}

\begin{equation}
wr_t = -100 * \frac{(hh_t - close_t)}{(hh_t - ll_t)}
\end{equation}    

\textbf{Awesome Oscillator} -  The Awesome Oscillator is an indicator used to measure market momentum. AO calculates the difference of a 34 Period and 5 Period Simple Moving Averages. The Simple Moving Averages that are used are not calculated using closing price but rather each bar's midpoints. AO is generally used to affirm trends or to anticipate possible reversals. 

\begin{equation}    
mp_t = \frac{(high + low)}{2}
\end{equation}
\begin{equation} 
ao_t = MA(mp_t,s) - MA(mp_t,2)
\end{equation}


\printbibliography


\end{document}